\documentclass[lettersize,journal]{IEEEtran}
\usepackage{amsmath,amsfonts}
\usepackage{algorithm}
\usepackage{array}
\usepackage[caption=false,font=normalsize,labelfont=sf,textfont=sf]{subfig}
\usepackage{textcomp}
\usepackage{stfloats}
\usepackage{url}
\usepackage{verbatim}
\usepackage{graphicx}
\usepackage{cite}

\usepackage{graphicx} 
\usepackage{multirow}
\usepackage[table]{xcolor}
\usepackage{hyperref}
\usepackage{booktabs}
\usepackage{enumitem}

\usepackage{algpseudocode}

\usepackage{times}
\usepackage{epsfig}
\usepackage{xspace}

\newcommand{\ourmodel}{FeTrIL\@\xspace}
\newcommand{\ourmodelNospace}{FeTrIL}

\newcommand{\ourmodelone}{FeTrIL$^1$~}

\newcommand\scalemath[2]{\scalebox{#1}{\mbox{\ensuremath{\displaystyle #2}}}}
\definecolor{armygreen}{rgb}{0.29, 0.33, 0.13}

\begin{document}

\title{FeTrIL++: Feature Translation for Exemplar-Free Class-Incremental Learning with Hill-Climbing}

\author{
    Eduard Hogea$^{3}$       \\{\tt\small eduard.hogea00@$\{$e-uvt$\}$.ro} \and 
    Adrian Popescu$^{1}$     \\{\tt\small adrian.popescu@cea.fr}          \and 
    Darian Onchis$^{3}$       \\{\tt\small darian.onchis@$\{$e-uvt$\}$.ro} \and 
    Grégoire Petit$^{1,2}$   \\{\tt\small gregoire.petit@$\{$cea,enpc$\}$.fr} \and 
    $^1$Université Paris-Saclay, CEA, List, F-91120, Palaiseau, France\\
    $^2$LIGM, Ecole des Ponts, Univ Gustave Eiffel, CNRS, Marne-la-Vallée, France\\
    $^3$West University of Timisoara, Timisoara, Romania
}

\author{
Eduard Hogea\textsuperscript{3},
Adrian Popescu\textsuperscript{1}, 
Darian Onchis\textsuperscript{3},
Grégoire Petit\textsuperscript{1,2}, 

 \textsuperscript{1}Université Paris-Saclay, CEA, LIST, F-91120, Palaiseau, France\\
 \textsuperscript{2}LIGM, Ecole des Ponts, Univ Gustave Eiffel, CNRS, Marne-la-Vallée, France\\
 \textsuperscript{3}West University of Timisoara, Timisoara, Romania\\
{\tt\small \{eduard.hogea00, darian.onchis\}@e-uvt.ro}\\
{\tt\small \{gregoire.petit, adrian.popescu\}@cea.fr}
}

\markboth{Journal of \LaTeX\ Class Files,~Vol.~14, No.~8, August~2021}%
{Shell \MakeLowercase{\textit{et al.}}: A Sample Article Using IEEEtran.cls for IEEE Journals}

\IEEEpubid{0000--0000/00\$00.00~\copyright~2021 IEEE}

\maketitle

\begin{abstract}
Exemplar-free class-incremental learning (EFCIL) poses significant challenges, primarily due to catastrophic forgetting, necessitating a delicate balance between stability and plasticity to accurately recognize both new and previous classes. Traditional EFCIL approaches typically skew towards either model plasticity through successive fine-tuning or stability by employing a fixed feature extractor beyond the initial incremental state. Building upon the foundational \ourmodel framework, our research extends into novel experimental domains to examine the efficacy of various oversampling techniques and dynamic optimization strategies across multiple challenging datasets and incremental settings. We specifically explore how oversampling impacts accuracy relative to feature availability and how different optimization methodologies, including dynamic recalibration and feature pool diversification, influence incremental learning outcomes.

The results from these comprehensive experiments, conducted on CIFAR100, Tiny-ImageNet, and an ImageNet-Subset, underscore the superior performance of \ourmodel in balancing accuracy for both new and past classes against ten contemporary methods. Notably, our extensions reveal the nuanced impacts of oversampling and optimization on EFCIL, contributing to a more refined understanding of feature-space manipulation for class incremental learning. \ourmodel and its extended analysis in this paper \ourmodelNospace$^{++}$ pave the way for more adaptable and efficient EFCIL methodologies, promising significant improvements in handling catastrophic forgetting without the need for exemplars. 
\end{abstract}


\section{Introduction}
\label{sec:introduction}
In this paper, we introduce several algorithmic developments and refinements of \ourmodel, as introduced in Petit et al. (2023)\cite{petit2023fetril}. We delve deeper into the capabilities of this recently proposed Enhanced Feature Consistency Incremental Learning (EFCIL) method through comprehensive experimentation. \ourmodel uniquely combines a frozen feature extractor with a pseudo-feature generator, leveraging geometric translation to maintain a robust representation of both new and past classes within the feature space. To further validate and refine our approach, we have embarked on a series of experiments to explore the impacts of oversampling and optimization techniques on incremental learning performance.

Our analysis reveals that the utility of oversampling is contingent on the feature density across classes. Specifically, in scenarios where the feature count per image is relatively low, oversampling can significantly boost accuracy. Conversely, when the feature count is inherently high, oversampling may lead to diminished returns. This effect is pronounced across different datasets such as CIFAR100 and Tiny-ImageNet, each comprising 500 images, versus ImageNet-Subset with 1500 images.
These findings underscore the nuanced relationship between feature availability and incremental learning efficacy, as documented in Table \ref{tab:combined_difference_accuracy}.

Further experimentation with a dynamic recalibration technique demonstrates marked benefits for large and diverse datasets. By more accurately mirroring the shifting data distribution across learning states, this method ensures the pseudo-features remain closely aligned with actual feature distributions, a crucial factor for sustaining accuracy amidst the addition of new classes. Comparatively, optimization methods, applied to the initial pseudo-features generated from the geometric translation, which draw features from an expanded pool including multiple new classes, exhibit performance enhancements by enriching the pseudo-feature composition.

A nuanced observation from our study involves the potential drawbacks of optimization methods that apply feature replacements where the feature pool lacks sufficient diversity, leading to a repetition of features in the optimized set. Our exploration of various optimization strategies, ranging from single-feature selection in to the diverse feature pooling and dynamic recalibration, illuminates the critical balance between feature diversity and optimization efficacy in enhancing model performance.

This expanded investigation not only reaffirms the robustness of the \ourmodel framework but also illuminates the intricate dynamics at play in feature-based incremental learning. By integrating these nuanced findings, we are poised to further refine \ourmodel's architecture and optimization techniques, driving forward the frontier of EFCIL research. As it was the case in the original work, we run experiments with a standard EFCIL setting~\cite{hou2019_lucir,zhu2021class,zhu2021pass}, which consists of a larger initial state, followed by smaller states which include the same number of classes. \\
\\
Results reaffirm that the proposed approach has better behavior compared to ten existing methods, including very recent ones.

\section{Related Work}
\label{sec:related}

Recent approaches in exemplar-free class-incremental learning (EFCIL) have explored various strategies to enhance model performance by using past class prototypes in combination with distillation techniques. For example, PASS~\cite{zhu2021pass} uses prototype augmentation to improve class discrimination in incremental states. IL2A~\cite{zhu2021class} generates features for past classes by utilizing class distribution information, which poses scalability challenges due to the need to store a covariance matrix for each class. Similarly, SSRE~\cite{zhu2022self} employs a prototype selection mechanism to distinguish between past and new classes more effectively.

\ourmodel and its advanced version, \ourmodel$^{++}$, adopt a strategy of using class prototypes to preserve information about past classes. A key feature of both models is the early freezing of the feature extractor, which distinguishes them from other methods that incorporate prototypes within a knowledge distillation process. This design choice simplifies the model and addresses the limitations found in distillation-based methods, especially when applied to large-scale datasets.

\ourmodel maintains only the centroids of past classes in memory, leveraging a geometric translation of features from new classes to generate pseudo-features. This approach provides a straightforward and efficient way to simulate the presence of past classes without requiring extensive data storage.

\ourmodel$^{++}$ builds upon this by additionally storing the diagonal of the covariance matrix for each past class, along with the centroids. This enhancement allows for a more nuanced representation of class variability, improving the model's ability to generate accurate pseudo-features for past classes. The inclusion of the diagonal of the covariance matrix enables \ourmodel$^{++}$ to approximate class distributions more precisely without the overhead of storing a full covariance matrix, as is necessary in IL2A.

\section{Proposed Method}
\label{sec:method}

The goal of class-incremental learning (CIL) is to learn classes that are introduced over time, in a sequence of stages including an initial stage and several incremental stages. The challenge is to recognize both new classes and those learned in earlier stages without having access to the old data, a setting known as exemplar-free CIL. This approach is particularly challenging because it must overcome catastrophic forgetting—a significant decline in the model's ability to recognize previous classes—without the ability to review past data.

Our method is designed to navigate this challenge. It consists of three key components: a feature extractor that captures essential data characteristics, a generator that creates pseudo-features to represent past classes, and an external classifier that distinguishes between all classes learned up to that point. We freeze the feature extractor early on to maintain a consistent way of understanding the data throughout the learning process. Instead of storing images of past classes, we generate pseudo-features for them, which, although not perfect, are effective enough to allow for the recognition of these classes in conjunction with the newly learned ones.

The process involves taking the characteristics (features) of new class images and transforming them into representations that mimic those of past classes, using known information about all classes (prototypes). These representations, or pseudo-features which are then guided via the optimization function, along with the actual features of new classes, are then used to train a classifier that can identify both old and new classes.

\subsection{Generation of pseudo-features}
\label{subsec:gen}
The proposed method is defined as: \vspace{-2mm}

\begin{equation}
    {\hat{f^t}}(c_p) = f(c_n) + \mu(C_p) - \mu(C_n)
\label{eq:generator}
\end{equation}
with: 
$C_p$ - target past class for which pseudo-features are needed; 
$C_n$ - new class for which images $b$ are available;
$f(c_n)$ - features of a sample $c_n$ of class $C_n$ extracted with $\mathcal{F}$;
$\mu(C_p), \mu(C_n)$ - mean features of classes $C_p$ and $C_n$ extracted with $\mathcal{F}$;
${\hat{f^t}}(c_p)$ - pseudo-feature vector of a pseudo-sample $c_p$ of class $C_p$ produced in the $t^{th}$ incremental state.

Eq.~\ref{eq:generator} translates the value of each dimension with the difference between the values of the corresponding dimension of $\mu(C_p)$ and $\mu(C_n)$. 
It creates a pseudo-feature vector situated in the region of the representation space associated to target class $C_p$ based on actual features of a new class $f(C_n)$.
The computational cost of generation is very small since it only involves additions and subtractions. 
$\mu(C_p)$ is needed to drive the geometric translation toward a region of the representation space which is relevant for $C_p$.
Centroids are computed when classes occur for the first time and then stored.
Their reuse is possible because $\mathcal{F}$ is fixed after the initial step and its associated features do not evolve. To accommodate multiple new classes in pseudo-feature generation, the formula is adjusted as follows:

\begin{equation}
    {\hat{f^t}}(c_p) = \bigoplus_{i=1}^{N} \left( f(c_{ni}) + \mu(C_p) - \mu(C_{ni}) \right)
\label{eq:multi_concat_generator}
\end{equation}

with:
\begin{itemize}
    \item $C_{ni}$ - each of the $N$ new classes for which features are available;
    \item $N$ - the number of new classes considered.
\end{itemize}
Equation~\ref{eq:multi_concat_generator} extends the original concept to concatenate features from $N$ new classes, directly integrating their modified representations to form an extended pseudo-feature vector for $C_p$. The resultant high-dimensional feature vector for $C_p$ offers a richer, more nuanced representation, leveraging the diversity across multiple new classes.

\subsection{Selection of pseudo-features}
\label{subsec:selection}
Eq.~\ref{eq:generator} translates the features for a single sample. 
If each class is represented by $s$ samples, the generation process needs to be repeated $s$ times.
When CIL states include several classes $ C_n$, the $s$ pseudo-features of each class $C_p$ can be obtained using different strategies, depending on how features of new classes are used. 
To address scenarios with oversampling from multiple new classes and enhance the representation of past classes with richer pseudo-features, we extend our methodology to include various strategies. The following strategies outline our approach to pseudo-feature selection, accommodating both single and multiple new class scenarios:

\begin{itemize}[nosep,leftmargin=*]
    \item \ourmodelNospace$^k$: $s$ features are transferred from the $k^{th}$ similar new class of each past class $C_p$. Similarities between the target $C_p$ and the $C_n$ available in the current state are computed using the cosine similarity between the centroids of each pair of classes. Experiments are run with different values of $k$ to assess if a variable class similarity has a significant effect on EFCIL performance. Since translation is based on a single new class, the distribution of pseudo-features will be similar to that of features of $C_n$, but in the region of the representation space around $\mu(C_p)$.

    \item \ourmodelNospace$^{rand}$: $s$ features are randomly selected from all new classes. This strategy assesses whether a more diversified source of features from different $C_n$ produces an effective representation of class $C_p$. 

    \item \ourmodelNospace$^{herd}$: $s$ features are selected from any new class based on a herding algorithm~\cite{max2009_herding}. It assumes that sampling \cite{onchis19, sensor} should include features which produce a good approximation of the past class. Herding was introduced in exemplar-based CIL to obtain an accurate approximation of each class by using only a few samples~\cite{rebuffi2017_icarl} and its usefulness was later confirmed \cite{hou2019_lucir,wu2019_bic,slim2022dataset}. It is adapted here to obtain a good approximation of the sample distribution of $C_p$ with $s$ pseudo-features.

    \item \ourmodelNospace$^{m2}$: $s$ features are selected from the two closest classes from all the available new classes. This method generates pseudo-features for past class $C_p$ by concatenating translated features from $C_{n1}$ and $C_{n2}$. The resulting high-dimensional feature vector enhances class representation and aids Linear SVC in identifying discriminative features without unduly increasing the model's complexity.

    \item \ourmodelNospace$^{m3}$: extending the previous one, this method includes features from the three nearest new classes, $C_{n1}$, $C_{n2}$, and $C_{n3}$, to generate pseudo-features for $C_p$. Concatenation of translated features from these classes creates a more complex and rich representation, leveraging the diversity of multiple new classes to improve model robustness and generalization.
\end{itemize}

To enhance the creation of pseudo-features, we introduce a hill climbing optimization process that minimizes the Euclidean distance between the diagonal elements of the covariance matrix of generated pseudo-features, $\hat{f^{t}}(C_{p})$, and the 'true diagonal',  $\mathbf{d}\_{\text{actual}}$. The 'true diagonal' represents the diagonal of the actual features' covariance matrix for class $C_p$, setting the variance profile target for pseudo-features. This optimization, achieved by selectively replacing features from $\hat{f^{t}}(C{p})$ with alternatives from an extended pool $X_{\text{extended}}$, is governed by a maximum number of iterations and a patience threshold. It aims to refine the feature alignment method to produce pseudo-features that more closely replicate the original class features' variance, thereby enhancing their representativeness and discrimination capability.

The optimization objective is concisely captured as:

\[
\min_{\hat{f}^{t}(C_p)} \lVert \mathbf{D}_{\text{pseudo}} - \mathbf{d}_{\text{actual}} \rVert_2
\]

where \(\lVert \cdot \rVert_2\) denotes the Euclidean norm. This aims to align the variance of pseudo-features with that of actual features, enhancing the methodological approach to feature pool creation and replacement and making the pseudo-features generated resemble the original distribution of the features. The methods introduced that make use of the hill climbing optimization are the following:

\begin{itemize}
    \item \ourmodelNospace$^{single}$: this method initiates with selecting $s$ features from $\hat{f}^{t}(C_{p})$ as the initial pseudo-features. The remaining of $\hat{f}^{t}(C_{p})$ constitute the feature pool. Optimization seeks to replace features in the initial set with those from the feature pool to achieve the minimization of the objective function.

    \item \ourmodelNospace$^{multi}$: $s$ features are chosen from $\hat{f}^{t}(C_{p})$, to form the initial pseudo-feature set. The feature pool, however, is expanded to include $s$ features from each of the $N-1$ new classes, denoted as $C_{n2}, C_{n3}, \ldots, C_{nN}$ (notice the omittment of $C_{n1}$, as it was used for the initial pseudo-feature set). This creates a diverse pool of features from across all new classes available in the current incremental state. The optimization process is then applied.

    \item  \ourmodelNospace$^{shift}$: An enhancement of the \ourmodelNospace$^{multi}$ method, this new method incorporates dynamic centroid recalibration for class $C_{p}$ after each potential feature replacement within the optimization process. Following the recalibration, features in the pseudo-feature set are adjusted to better align with this newly calculated centroid of $C_{p}$.

    \item \ourmodelNospace$^{m2_{opt}}$ and \ourmodelNospace$^{m3_{opt}}$: Both methods apply hill climbing optimization to the initial $s$ pseudo-features selected from $\hat{f}^{t}(C_{p})$. The feature pool for replacements includes the same features as those initially selected, focusing on optimizing this specific subset.

    \item \ourmodelNospace$^{M2_{opt}}$ and \ourmodelNospace$^{M3_{opt}}$: Variants of \ourmodelNospace$^{m2}$ and \ourmodelNospace$^{m3}$ respectively, these methods extend the feature pool with $s$ features for all available classes. Hill climbing optimization is applied afterwards.

\end{itemize}

\begin{algorithm}
\caption{Optimize Pseudo-Features via Hill Climbing}\label{alg:hill_climbing}
\begin{algorithmic}[1]
\State \textbf{Input:} $\hat{f}^{t}(C_{p})$ as $X_{\text{tmp}}$, $X_{\text{ext}}$, $\mathbf{d}_{\text{actual}}$, $\mathbf{D}_{\text{pseudo}}$, $maxIter$, $replaceCnt$, $patience$
\State \textbf{Output:} Optimized $X_{\text{tmp}}$

\State $currDist \gets$ computeDistance($\mathbf{d}_{\text{actual}}$, $\mathbf{D}_{\text{pseudo}}$)
\State $noImprove \gets 0$
\State $iter \gets 0$

\While{$iter < maxIter$ \textbf{and} $noImprove < patience$}
    \State $idxReplace \gets$ rnd(len($X_{\text{tmp}}$), $replaceCnt$, False) \Comment{Random indices for replacement}
    \State $idxExt \gets$ rnd(len($X_{\text{ext}}$), $replaceCnt$, False) \Comment{Indices from extended set for substitution}
    
    \State $propFeatures \gets X_{\text{tmp}}$
    \State $propFeatures[idxReplace] \gets X_{\text{ext}}[idxExt]$ \Comment{Propose new feature set}
    
    \State $newDist \gets$ computeDistance($propFeatures$, $\mathbf{d}_{\text{tgt}}$)
    
    \If{$newDist < currDist$}
        \State $X_{\text{tmp}} \gets propFeatures$ \Comment{Adopt improved set}
        \State $currDist \gets newDist$
        \State $noImprove \gets 0$
    \Else
        \State $noImprove \gets noImprove + 1$
    \EndIf
    \State $iter \gets iter + 1$
\EndWhile

\State \textbf{return} $X_{\text{tmp}}$
\end{algorithmic}
\end{algorithm}

The comparison of these different strategies will allow us to determine whether the geometric translation of features is prevalent, or if a particular configuration of the features around the centroid of the target past class is needed. 

\subsection{Linear classification layer training}
\label{subsec:layer}
We assume that the CIL process is in the $t^{th}$ CIL state, which includes $P$ past classes and $N$ new classes.
The combination of the feature generator (Subsection~\ref{subsec:gen}) and selection (Subsection~\ref{subsec:selection}) provides a set ${\hat{f^t}}(C_p)$ of $s$ pseudo-features for each class $C_p$. 
The objective is to train a linear classifier for all $P+N$ seen classes, which takes pseudo features of past classes and actual features of new classes as inputs.
This linear layer is defined as:
\begin{equation}
\scalemath{0.9}{
    \mathcal{W}^t = \{w^t(C_1),..., w^t(C_P), w^t(C_{P+1}),..., w^t(C_{P+N})\}
}
    \label{eq:linear}
\end{equation}
with: $w^t$ - the weight of known classes in the $t^{th}$ CIL state. 

$\mathcal{W}^t$ can be implemented using different classifiers, such as LinearSVCs as external ones or fully-connected layer to enable end-to-end training.

\section{Results}
\label{subsec:results}

\begin{table*}[t]
\begin{center}
\resizebox{0.95\textwidth}{!}{
\begin{tabular}{@{\kern0.5em}llccccccccccccccccccc@{\kern0.5em}}
        \toprule
        \multirow{2}{*}{\textbf{CIL Method}}
            & \multicolumn{4}{c}{CIFAR-100}
            && \multicolumn{4}{c}{TinyImageNet}
            && \multicolumn{4}{c}{ImageNet-Subset}
            && \multicolumn{3}{c}{ImageNet}
        \\ \cmidrule(lr){2-5} \cmidrule(lr){7-10} \cmidrule(l){12-15} \cmidrule{17-19}  
        
            & \multicolumn{1}{c}{\textit{T}=5}
            & \multicolumn{1}{c}{\textit{T}=10}
            & \multicolumn{1}{c}{\textit{T}=20}
            & \multicolumn{1}{c}{\textit{T}=60}
            &
            & \multicolumn{1}{c}{\textit{T}=5}
            & \multicolumn{1}{c}{\textit{T}=10}
            & \multicolumn{1}{c}{\textit{T}=20}
            & \multicolumn{1}{c}{\textit{T}=100}
            &
            & \multicolumn{1}{c}{\textit{T}=5}
            & \multicolumn{1}{c}{\textit{T}=10}
            & \multicolumn{1}{c}{\textit{T}=20}
            & \multicolumn{1}{c}{\textit{T}=60}
            &
            & \multicolumn{1}{c}{\textit{T}=5}
            & \multicolumn{1}{c}{\textit{T}=10}
            & \multicolumn{1}{c}{\textit{T}=20}
            
        \\ \midrule

EWC$^*$~\cite{kirkpatrick2017overcoming} \small (PNAS'17)    & 24.5 & 21.2 & 15.9 & x && 18.8 & 15.8 & 12.4 & x && -    & 20.4 & - & x  && - & - & - \\
LwF-MC$^*$~\cite{rebuffi2017_icarl} \small (CVPR'17) & 45.9 & 27.4 & 20.1 & x && 29.1 & 23.1 & 17.4 & x && -    & 31.2 & - & x  && - & - & - \\    
LUCIR \small (CVPR'19)      & 51.2 & 41.1 & 25.2 & x && 41.7 & 28.1 & 18.9 & x && 56.8 & 41.4 & 28.5  & x && 47.4 & 37.2 & 26.6 \\
MUC$^*$~\cite{liu2020more} \small (ECCV'20)    & 49.4 & 30.2 & 21.3 & x && 32.6 & 26.6 & 21.9 & x && -    & 35.1 & - & x && - & - & - \\   
SDC$^*$~\cite{sdc_2020} \small (CVPR'20)    & 56.8 & 57.0 & 58.9 & x &&   -  & -    &-  & x  && -    & 61.2 & - & x && - & - & - \\   
ABD$^*$~\cite{smith2021always} \small (ICCV'21)    & 63.8 & 62.5 & 57.4 & x &&   -  & -    &- & x    && -    & - & - & x && - & - & - \\ 
PASS$^*$~\cite{zhu2021pass} \small (CVPR'21)   & 63.5 & 61.8 & 58.1 & x && 49.6 & 47.3 & 42.1 & x && 64.4   & 61.8 & {51.3} & x && - & - & - \\ 
IL2A$^*$~\cite{zhu2021class} \small (NeurIPS'21)& 66.0 & 60.3 & 57.9 & x && 47.3 & 44.7 & 40.0  & x && - & - & - & x  && - & - & - \\
BSIL$^*$~\cite{jodelet2021balanced} \small (ICANN'21)     & \underline{67.2} & 64.6 & -    & x && \textbf{53.8} & 51.4 & 42.6 & x && \underline{71.8} & 67.2 & -     & x && - & - & - \\ 
SSRE$^*$~\cite{zhu2022self} \small (CVPR'22)   & 65.9 & 65.0 & 61.7 & x && 50.4 & 48.9 & 48.2 & x && -    & 67.7 &  - & x && - & - & - \\ 
\hline
DeeSIL~\cite{belouadah2018_deesil} \small (ECCVW'18)    & 60.0 & 50.6 & 38.1 & x && 49.8 & 43.9 & 34.1 & x &&  {67.9} & 60.1 & 50.5 & x && 61.9 & 54.6 & 45.8 \\

DSLDA~\cite{hayes2020_deepslda} \small (CVPRW'20)   & 64.0 & 63.8 & 60.8 & 60.5 && 53.1 & \textbf{52.9} & \textbf{52.8} & 52.6 && 71.3  & \textbf{71.2} &  \textbf{71.0} & 70.8 && 64.0 & 63.8 & 63.6 \\

\ourmodelone  & 66.9 & \underline{65.8} & \underline{62.6} &  && \underline{53.3} & \underline{52.5} & 51.6 &  && \underline{71.8} & \underline{70.8} &67.8 &  &&  &  &  \\

\ourmodel$^{shift}$  & \textbf{67.3} & \textbf{66.4} & \textbf{62.9} & && \textbf{53.8} & \textbf{52.9} & \underline{52.1} &  && \textbf{72.3} & \textbf{71.2} & \underline{68.0} &  &&  &  &  \\

\hline
\end{tabular}
}
\end{center}
\vspace{-2mm}
	\caption{Average top-1 incremental accuracy in EFCIL with different numbers of incremental steps.
	\ourmodelone results are reported with pseudo-features translated from the most similar new class.
	"-" cells indicate that results were not available (see supp. material for details). "x" cells indicate that the configuration is impossible for that method. 
	\textbf{Best results - in bold}, \underline{second best - underlined}.\vspace{-4.3mm}}
\label{tab:main_results}
\end{table*}

\subsection{Method analysis}
\label{subsec:analysis}
\vspace{-1mm}

We present an analysis of: the impact of oversampling and the effect of the hill climbing optimization.


\textbf{Oversampling methods comparison.} The oversampling scenarios proved to increase the accuracy in cases where the number of features was smaller, and decrease it when the number was already high. Even if for each image the number of available features is the same, CIFAR100 and Tiny-ImageNet have 500 images each, while ImageNet-Subset has 1500. This directly affects the total number of features across the classes, and the results are showed in Table \ref{tab:combined_difference_accuracy}.


\begin{table}[!h]
\centering
\caption{Difference in Accuracy for Oversampling Methods Compared to \ourmodel Across Datasets at Each Incremental Stage. The methods work as intended when the total features available per class is lower.}
\label{tab:combined_difference_accuracy}
\resizebox{0.48\textwidth}{!}{%
\begin{tabular}{lcccccccc}
\hline
\multirow{2}{*}{Dataset \& CIL Method} & \multicolumn{2}{c}{t = 5} & \multicolumn{2}{c}{t = 10} & \multicolumn{2}{c}{t = 20} & \multicolumn{2}{c}{Avg. Acc. Change} \\
                                       & Top 1       & Top 5       & Top 1        & Top 5       & Top 1        & Top 5       & Top 1           & Top 5           \\ \hline
CIFAR100 \ourmodelNospace              & x           & x           & x            & x           & x            & x           & x               & x               \\
CIFAR100 \ourmodelNospace$^{single}$   & -1.1        & -0.4        & -1.0         & -0.6        & -1.2         & -0.5        & -1.1            & -0.5            \\
CIFAR100 \ourmodelNospace$^{m2}$       & +0.4        & +0.2        & +0.4         & +0.2        & +0.3         & +0.1        & +0.4            & +0.2            \\
CIFAR100 \ourmodelNospace$^{m3}$       & +0.4        & +0.2        & +0.3         & +0.2        & +0.3         & -0.1        & +0.3            & +0.1            \\
Tiny-ImageNet \ourmodelNospace         & x           & x           & x            & x           & x            & x           & x               & x               \\
Tiny-ImageNet \ourmodelNospace$^{single}$ & -1.2    & -1.1        & -1.1         & -1.0        & -1.0         & -1.0        & -1.1            & -1.0            \\
Tiny-ImageNet \ourmodelNospace$^{m2}$  & +0.2        & -0.2        & +0.2         & 0           & +0.2         & -0.1        & +0.2            & -0.1            \\
Tiny-ImageNet \ourmodelNospace$^{m3}$  & +0.3        & -0.3        & +0.1         & -0.1        & +0.1         & -0.2        & +0.2            & -0.2            \\
ImageNet-Subset \ourmodelNospace       & x           & x           & x            & x           & x            & x           & x               & x               \\
ImageNet-Subset \ourmodelNospace$^{single}$ & -0.1  & +0.0        & -0.0         & -0.1        & -0.1         & +0.0        & -0.1            & -0.0            \\
ImageNet-Subset \ourmodelNospace$^{m2}$ & -0.0      & -0.2        & -0.2         & -0.2        & -0.2         & -0.3        & -0.1            & -0.2            \\
ImageNet-Subset \ourmodelNospace$^{m3}$ & -0.2      & -0.2        & -0.5         & -0.2        & -0.4         & -0.4        & -0.4            & -0.3            \\ \hline
\end{tabular}%
}
\end{table}

\textbf{Optimization effects.} The experimental results indicate that dynamic recalibration, as implemented in the \ourmodelNospace$^{shift}$ method, is particularly beneficial for large and diverse datasets. This advantage is likely due to its ability to more accurately capture the evolving data distribution across incremental learning states. Methods that incorporate features from multiple new classes, such as \ourmodelNospace$^{multi}$, \ourmodelNospace$^{M2_{opt}}$, and \ourmodelNospace$^{M3_{opt}}$, also demonstrate improvements in performance by enriching the pseudo-feature set. However, the recalibration step in \ourmodelNospace$^{shift}$ provides a critical advantage, ensuring that the generated pseudo-features remain closely aligned with the actual feature distribution. This alignment is crucial for maintaining and enhancing model accuracy in the face of new class additions, especially in contexts characterized by significant dataset size and diversity (Tables \ref{tab:res1}, \ref{tab:res2}, \ref{tab:res3}).

An interesting study is also performed in methods such as \ourmodelNospace$^{m2_{opt}}$ and \ourmodelNospace$^{m3_{opt}}$, where it was shown that the optimization method can be also detrimental when the feature pool doesn't contain enough diversity. In those methods, feature pool for replacements was identical to the initial pseudo-features, essentially repeating some features in the resulted optimized set.

\begin{table}[!h]
\caption{Results for CIFAR100, with the first, non-incremental state containing half of the classes (except for t = 20, where 40 classes were used). Feature extractor: ResNet18(LUCIR+AugMix). Avg. Acc. Change is calculated as the average change  to the original accuracy for each method.}
\label{tab:res1}
\resizebox{\linewidth}{!}{%
\footnotesize
\begin{tabular}{ccccccccc}
\hline
\multirow{3}{*}{CIL method}   & \multicolumn{2}{c}{t = 5} & \multicolumn{2}{c}{t = 10} & \multicolumn{2}{c}{t = 20} & \multicolumn{2}{c}{Avg. Acc. Change} \\
                              & Top 1       & Top 5       & Top 1        & Top 5       & Top 1        & Top 5       & Top 1           & Top 5           \\ \hline
\ourmodelNospace &66.85 &86.22 &65.81 &85.80 &62.59 &84.53 & - & - \\
\ourmodelNospace$^{single}$ &65.77 &85.83 &64.85 &85.19 &61.43 &84.03 & -1.07 & -0.50 \\
\ourmodelNospace$^{multi}$ &67.21 &86.61 &66.34 &86.24 &62.86 &84.64 & 0.39 & 0.31 \\
\ourmodelNospace$^{shift}$ &67.27 &86.71 &66.40 &86.28 &62.96 &84.81 & 0.46 & 0.42 \\
\ourmodelNospace$^{m2_{opt}}$ &66.93 &86.35 &66.01 &85.99 &62.78 &84.51 & 0.16 & 0.10 \\
\ourmodelNospace$^{m3_{opt}}$ &67.14 &86.36 &66.03 &85.99 &62.74 &84.47 & 0.22 & 0.09 \\
\ourmodelNospace$^{m2}$ &67.22 &86.44 &66.18 &86.02 &62.87 &84.61 & 0.34 & 0.17 \\
\ourmodelNospace$^{m3}$ &67.28 &86.38 &66.09 &85.99 &62.85 &84.45 & 0.32 & 0.09 \\
\ourmodelNospace$^{M2_{opt}}$&67.42 &86.86 &66.36 &86.32 &63.02 &84.72 & 0.52 & 0.45 \\
\ourmodelNospace$^{M3_{opt}}$&67.50 &86.84 &66.27 &86.18 &- &- & 0.55 & 0.50 \\
\ourmodelNospace$^{upper}$ &69.23 &87.97 &69.04 &87.82 &67.07 &87.61 & - & - \\ \hline
\end{tabular}%
}
\end{table}

\begin{table}[!h]
\caption{Results for Tiny-ImageNet, with the first, non-incremental state containing half of the classes. Feature extractor: ResNet18(LUCIR+AugMix). Avg. Acc. Change is calculated as the average change to the original accuracy for each method.}
\label{tab:res2}
\resizebox{\linewidth}{!}{%
\footnotesize
\begin{tabular}{ccccccccc}
\hline
\multirow{3}{*}{CIL method} & \multicolumn{2}{c}{t = 5} & \multicolumn{2}{c}{t = 10} & \multicolumn{2}{c}{t = 20} & \multicolumn{2}{c}{Avg. Acc. Change} \\
                            & Top 1       & Top 5       & Top 1        & Top 5       & Top 1        & Top 5       & Top 1           & Top 5           \\ \hline
\ourmodelNospace & 53.33 & 73.83 & 52.49 & 73.29 & 51.62 & 72.83 & - & - \\
\ourmodelNospace$^{single}$ & 52.15 & 72.68 & 51.39 & 72.24 & 50.60 & 71.80 & -1.10 & -1.08 \\
\ourmodelNospace$^{multi}$ & 53.32 & 73.86 & 52.66 & 73.48 & 51.90 & 73.08 & +0.15 & +0.16 \\
\ourmodelNospace$^{shift}$ & 53.77 & 74.22 & 52.94 & 73.73 & 52.05 & 73.36 & +0.44 & +0.45 \\
\ourmodelNospace$^{m2_{opt}}$ & 53.39 & 73.38 & 52.39 & 73.02 & 51.56 & 72.64 & -0.03 & -0.30 \\
\ourmodelNospace$^{m3_{opt}}$ & 53.40 & 73.39 & 52.44 & 73.09 & 51.49 & 72.56 & -0.04 & -0.30 \\
\ourmodelNospace$^{m2}$ & 53.50 & 73.66 & 52.65 & 73.30 & 51.78 & 72.78 & +0.16 & -0.07 \\
\ourmodelNospace$^{m3}$ & 53.62 & 73.56 & 52.63 & 73.19 & 51.71 & 72.68 & +0.17 & -0.17 \\
\ourmodelNospace$^{M2_{opt}}$ & 53.67 & 74.19 & 52.88 & 73.73 & 52.01 & 73.08 & +0.37 & +0.35 \\
\ourmodelNospace$^{M3_{opt}}$ & 53.97 & 74.06 & 52.91 & 73.63 & 51.89 & 72.90 & +0.44 & +0.21 \\
\ourmodelNospace$^{upper}$ & 55.27 & 75.48 & 55.19 & 75.26 & 55.11 & 75.15 & - & - \\ \hline
\end{tabular}%
}
\end{table}

\begin{table}[!h]
\caption{Results for ImageNet-Subset. For All the Tests, the First, Non-Incremental, State Contains Half of the Classes, except for t = 20 where 40 classes have been used for the initial model. Feature Extractor Is ResNet18(LUCIR+AugMix)}
\label{tab:res3}
\resizebox{\linewidth}{!}{%
\footnotesize
\begin{tabular}{ccccccccc}
\hline
\multirow{3}{*}{CIL method} & \multicolumn{2}{c}{t = 5} & \multicolumn{2}{c}{t = 10} & \multicolumn{2}{c}{t = 20} & \multicolumn{2}{c}{Avg. Acc. Change} \\
                            & Top 1 & Top 5 & Top 1 & Top 5 & Top 1 & Top 5 & Top 1 & Top 5 \\ \hline
\ourmodelNospace & 71.81 & 86.68 & 70.84 & 86.52 & 67.79 & 84.37 & - & - \\
\ourmodelNospace$^{single}$ & 71.76 & 86.71 & 70.82 & 86.38 & 67.69 & 84.39 & -0.06 & -0.03 \\
\ourmodelNospace$^{multi}$ & 71.98 & 86.61 & 71.02 & 86.49 & 67.88 & 84.46 & 0.15 & -0.00 \\
\ourmodelNospace$^{shift}$ & 72.27 & 86.88 & 71.23 & 86.66 & 68.04 & 84.57 & 0.37 & 0.18 \\
\ourmodelNospace$^{m2_{opt}}$ & 71.64 & 86.41 & 70.55 & 86.13 & 67.55 & 84.10 & -0.23 & -0.31 \\
\ourmodelNospace$^{m3_{opt}}$ & 71.59 & 86.41 & 70.51 & 86.17 & 67.52 & 83.96 & -0.27 & -0.34 \\
\ourmodelNospace$^{m2}$ & 71.78 & 86.48 & 70.60 & 86.29 & 67.62 & 84.10 & -0.15 & -0.23 \\
\ourmodelNospace$^{m3}$ & 71.64 & 86.47 & 70.38 & 86.29 & 67.37 & 83.98 & -0.35 & -0.28 \\
\ourmodelNospace$^{M2_{opt}}$ & 71.90 & 86.49 & 70.84 & 86.53 & 67.64 & 84.26 & -0.02 & -0.10 \\
\ourmodelNospace$^{M3_{opt}}$ & 71.86 & 86.62 & 70.64 & 86.34 & - & - & -0.08 & -0.12 \\
\ourmodelNospace$^{upper}$ & 74.66 & 88.84 & 74.40 & 88.69 & 71.94 & 87.89 & - & - \\ \hline
\end{tabular}%
}
\end{table}


\section{Conclusion}
\label{sec:conclusions}
In this work, we have extended \ourmodel, a novel approach to exemplar-free class-incremental learning (EFCIL) that adeptly leverages a frozen feature extractor and a pseudo-feature generator for enhanced incremental learning. The pseudo-features, generated through geometric translation, offer a simple yet effective means to represent past classes, striking a delicate balance between stability and plasticity. This methodology not only showcases superior results compared to contemporary EFCIL approaches~\cite{hou2019_lucir,rebuffi2017_icarl,smith2021always,kumar2021_efficient,sdc_2020,zhu2021class,zhu2021pass,zhu2022self} but also exhibits significant advantages in terms of memory usage and computational speed, critical for deployment on edge devices with limited resources~\cite{hayes2022online,ravaglia2021tinyml}.

Through extensive experimentation, including the investigation of oversampling techniques and optimization effects across various challenging datasets, our extended analysis in \ourmodelNospace$^{++}$ demonstrates \ourmodel's robust performance, narrowing the gap between exemplar-based and exemplar-free methods. The outcomes of our work underscore the potency of straightforward strategies in addressing catastrophic forgetting within CIL contexts~\cite{masana2021_study,prabhu2020gdumb}, and prompt a reevaluation of the prevalent use of knowledge distillation among existing EFCIL solutions. Also, the proposed methods will be further tested in the context of online damage detection settings \cite{onchis20}.  

Looking forward, we acknowledge certain limitations that will guide our future endeavors. While \ourmodel effectively utilizes a frozen feature extractor, enhancing the representation of past over new classes, we aim to explore integrations of the pseudo-feature generation with selective fine-tuning strategies to elevate overall performance and refine the equilibrium between stability and plasticity. Moreover, the generation of pseudo-features that more closely mimic the original class features remains a pivotal area for enhancement. Investigating methodologies for producing more accurate feature representations, potentially through the exploitation of initial feature distributions, will be a primary focus. Additionally, our future work will seek to optimize the selection of pseudo-features by devising methods to exclude outliers, thereby ensuring a more authentic and cohesive representation of past classes within the feature space.


{\small
\bibliographystyle{ieee_fullname}
\bibliography{egbib}
}

\end{document}